\documentclass{article}

\PassOptionsToPackage{numbers, compress}{natbib}

     \usepackage[preprint]{neurips_2019}

\usepackage[utf8]{inputenc} 
\usepackage[T1]{fontenc}    
\usepackage{hyperref}      
\usepackage{url}            
\usepackage{booktabs}       
\usepackage{amsfonts}       
\usepackage{amsmath}
\usepackage{nicefrac}       
\usepackage{microtype}      
\usepackage{graphicx}
\usepackage{multirow}
\usepackage{natbib}

\title{Demand Forecasting from Spatiotemporal Data with Graph Networks and Temporal-Guided Embedding}

\author{
  Doyup Lee \thanks{Work done during an intership at Kakao Brain.} \\
  POSTECH\\
  \texttt{doyup.lee@postech.ac.kr} 
  \And
  Suehun Jung \thanks{Work done at Kakao Brain} \\
  Statice GmbH\\
  \texttt{sue@statice.io} 
  \And
  Yeongjae Cheon \\
  Kakao Brain \\
  \texttt{yeongjae.cheon@kakaobrain.com} 
  \And
  Dongil Kim\\
  Kakao Mobility \\
  \texttt{bob033@kakaomobility.com}
  \And
  Seungil You \\
  Kakao Mobility \\
  \texttt{sean.you@kakaomobility.com}
  }

\begin{document}

\maketitle

\begin{abstract}
  For accurate demand forecasting, previous approaches build complex models, use long-term histories of demand as input, and utilize external data sources.
  However, in this study, we propose Temporal-Guided Network (TGNet), which is an sample and efficient baseline model of short-term demand forecasting.
  Graph networks can extract complex spatiotemporal features of each region, and the features are invariant to the locational permutation of adjacent regions.
  Temporal-guided embedding can learn temporal contexts explicitly and capture temporally recurrent patterns instead of long-term demand histories.
  In the experiments on real-word datasets, our model shows competitive performances with other compared models.
  The forecasting performances are achieved without external data sources, and the number of trainable parameters is about 20 times smaller than a recent state-of-the-art model.
  Temporal-guided embedding shows interpretable visualization, which represents temporal contexts in training time series.
  Finally, we also show that our simple and powerful baseline is well generalized to minority samples with extremely large value, which are important in real-world situations and rarely appear in training time.
\end{abstract}

\section{Introduction}
On-demand ride hailing platforms, such as Uber, Didi, and Lyft, fulfill more than millions ride hailing requests per day and become the essential part of urban lift style.
Short-term demand forecasting is crucial in those platforms since dispatch system's efficiency can be improved by dynamic adjustment of the fare and relocation of idle drivers to area with high demand.

\begin{figure*}[t]
  \centering
  \includegraphics[width=14cm]{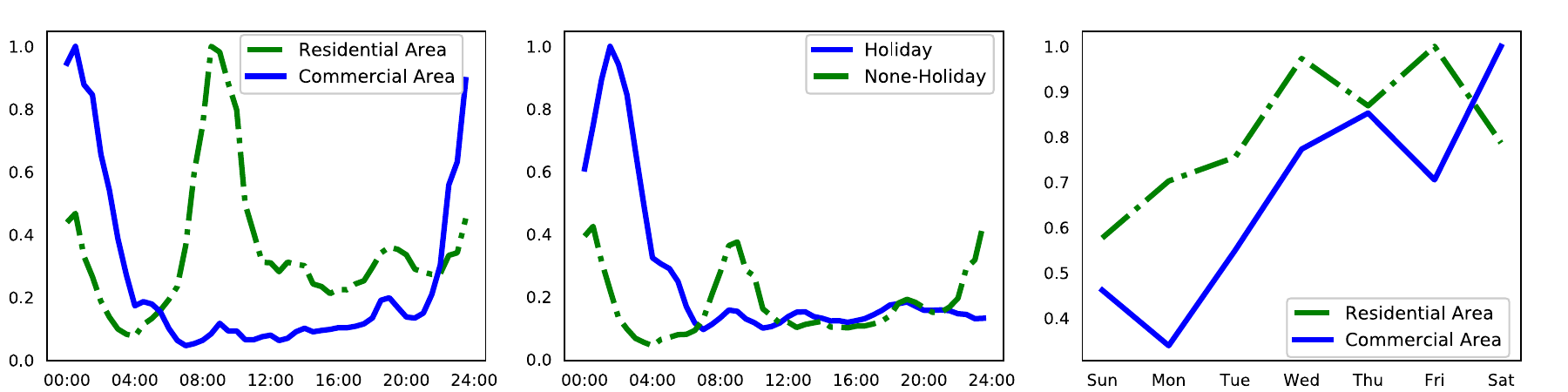}
  \caption{Ride-hailing Demands in Seoul: Daily patterns of pick-up requests in average (left) and on holiday (middle) according to time-of-day. Demand patterns are also different according to day-of-week (right). The scales are normalized.}
  \label{fig_data_stat}
\end{figure*}

A predictive model must consider spatial and temporal dependencies, and external influences for accurate demand forecasting \cite{zhang2017deep}.
Recent approaches show prominent performances, combining convolutional neural networks \cite{lecun1998gradient} and long-short term memories  \cite{hochreiter1997long} to extract spatial and temporal features respectively \cite{zhang2016dnn,yao2018deep,yao2018modeling}.
In temporal dependencies, another important issue is temporally recurrent patterns, because periodic and seasonal patterns are commonly appear in real-world time-series (Figure~\ref{fig_data_stat}).
Thus, long-term histories from days/weeks ago are also used as input to consider periodicity and seasonality in temporal patterns \cite{zhang2016dnn,zhang2017deep,yao2018modeling}

Recent studies continue to improve the accuracy of demand forecasting \cite{zhang2017deep,yao2018deep,yao2018modeling}.
However, we tackle that the studies focus on building complex models, using long-term history as input, depend on external data sources, and increasing the number of trainable parameters excessively. 
For example, STDN \cite{yao2018modeling} contains about 9.5 millions of parameters to learn 1,523 training samples with 10$\times$20 grid regions\footnote{ResNet-110 uses 1.7 millions of parameters to train ImageNet dataset \cite{he2016deep}.}

In this paper, we propose an efficient baseline model with graph networks and temporal-guided embedding, Temporal-Guided Network (TGNet).
TGNet has simple architecture with a stacks of graph networks and show competitive performances on real-world datasets.
However, TGNet has about 20 times smaller number of trainable parameters than STDN.
There are \textit{three} main differences in our modeling from previous approaches.

\textbf{Permutation-Invariant Operation in Graph Networks}
We show that permutation-invariant operation can be enough to extract spatial features of a region, both reducing the size of model and improving forecasting performances.
Convolution calculates different values according to the locational permutations of regions in each receptive field in general.
For example, when a subway station is adjacent to the target region, convolution consider the directionality of subway station (whether the subway station is in east or west from the region.
However, permutation-invariant operation in graph networks considers only the proximity of the subway station (whether it is adjacent to the target region or not).

\textbf{Spatiotemporal Feature Extraction}
For modeling spatiotemporal features in demand patterns, recent approaches extract spatial and temporal features separately  \cite{yao2018deep,yao2018modeling}.
After spatial features are extracted from demand pattern at each time, the extracted spatial features are used as the input of LSTM layer, which considers temporal dependencies between the spatial features in the order of time steps.
Instead of these hierarchical feature extraction, we extract complex spatiotemporal features at the same time.

\textbf{Temporal-Guided Embedding}
We propose temporal-guided embedding to effectively digest temporally recurrent patterns, such as periodicity and seasonality, instead of the input of days/weeks ago histories.
The demands of days/weeks ago  from target time makes a model directly refer past demand patterns in similar temporal contexts of forecasting target, but increase the number of trainable parameters and do not learn temporal contexts explicitly from training data.

When a person understands time series, she or he recognizes temporal information from time-of-day, day-of-week, and holiday information to learn temporal contexts such as weekday morning rush hours.
Thus, we use temporal information to learn the encoding of temporal contexts \textit{explicitly} and concatenate the learned encoding on input demand patterns.
Temporal-guided embedding is simple idea, but always gives significant performance gains in our experiment.
In addition, temporal-guided embedding has meaningful and interpretable visualization of temporal contexts in data.

In addition to aforementioned differences in modeling, our contributions contain efficient and powerful performances of TGNet on real-world datasets.
On large-scale dataset, we fail to train other compared models to the best of our efforts, but TGNet are easily trained only by increasing the number of hidden neurons in each layer.
Furthermore, TGNet shows better generalization on minority samples such as atypical event samples, which are important in practice and have extremely large demand volumes, but rarely appear in training time.

\section{Related Work} \label{5}
Many predictive models are used to learn complex spatiotemporal patterns in demand data.
ARIMA is used to predict future traffic condition and exploit temporal pattern in a data \cite{pan2012utilizing,moreira2013predicting}.
Latent space model \cite{deng2016latent} or k-nearest neighbor (kNN) \cite{cheng2018short} are applied to capture spatial correlation between adjacent regions for short-term traffic forecasting.
While these approaches show promising progress on traffic forecasting, they have a limited capability to capture complex spatiotemporal patterns.

Most of recent models adopte convolutional neural networks (CNNs) \cite{lecun1998gradient} and long short-term memory (LSTMs) \cite{hochreiter1997long} to extract spatial and temporal features respectively.
To turn the geographical data into a 2D image, a grid over a region is formed and a quantity of interest is assigned as pixel value.
For example, the traffic speed of each region is turned into 2D image, and the future traffic speeds are predicted \cite{ma2017learning}.
Then, feature maps are extracted by a stack of convolutional layers, considering local relationship between adjacent regions \cite{zhang2016dnn,zhang2017deep,yao2018deep,yao2018modeling}.
After spatial features of each time step are extracted by CNN, various models use LSTM layers to capture autoregressive sequential dependency and forecast traffic amounts and conditions \cite{zhao2017lstm,cheng2017deeptransport}, taxi demands \cite{ke2017short,zhou2018predicting,yao2018deep,yao2018modeling}, or traffic speeds \cite{yu2017spatiotemporal}.
We note that they do not extract spatiotemporal features at once, but extract spatial and temporal features separately.

Some approaches utilize long-term history of demand volumes from days/weeks ago to improve forecasting performances, because demand patterns often have temporally recurrent patterns according to time-of-day, day-of-week, and holiday or not.
Three convolutional models are used to extract features of temporal closeness, period, and seasonal trend from immediate past, days, and weeks ago samples from forecasting target \cite{zhang2016dnn,zhang2017deep}.
The extracted features are also used as the input of LSTM layer with periodically shifted attention mechanisms \cite{yao2018modeling}.
In this study, we does not use long-term histories, but learn temporal contexts of forecasting target time and conditional distribution on immediate past samples and target temporal contexts.

Ingesting external data sources that may related with future demand can also improve forecasting performances.
For example, meteorological, event information \cite{zhang2016dnn,zhang2017deep,yao2018deep}, or traffic (in/out) flows \cite{yao2018modeling} can be used to improve forecasting results.
However, the improvement is orthogonal to capture of complex spatiotemporal dependencies in input data.
Thus, we focus on building an efficient baseline model to learn complex spatiotemporal patterns and achieve competitive performances.
Various data are expected to be combined with our model in future work.

Recent studies successfully apply neural networks to graph data and graph networks extract hidden representations of each node from the messages of its neighborhood.
For example, GraphSAGE \cite{hamilton2017inductive} learns a function to generate embedding of node by sampling and aggregating from its neighborhood. 
Message passing neural networks (MPNNs) \cite{gilmer2017neural} define message/update functions and integrate many previous studies on graph domains \cite{duvenaud2015convolutional,li2015gated,battaglia2016interaction,kearnes2016molecular,schutt2017quantum,kipf2016semi}.
With the development of graph networks, we extract spatiotemporal features of demand patterns in region, after the demands in a city are transformed into graph data.


\section{Problem Setting of Demand Forecasting} \label{2}
\subsection{Spatiotemporal Demand Data}
In spatiotemporal modeling, different tessellations, such as grid \cite{ma2017learning}, hexagon \cite{ke2018hexagon}, or others \cite{davis2018taxi}, are used to divide the area of interest into non-overlapped regions.
We use Non-overlapped grid tessellation and split the area of whole city into $I\times J$ grid map $L=\left\{L_{1}, ..., L_{I \times J} \right\}$.
Non-overlapped time intervals are also defined as $\mathcal{I}=\left\{I_{1}, ..., I_{p}\right\}$.
When a demand log $r$ of user has time stamp and its location $(r.t, r.l)$, demand in time interval $I_{t}$ is defined by
\begin{equation} \label{eq_x_def}
\textbf{x}^{(t)}=\left\{x^{(t)}_{i}=\left| \left\{ r:r.t\in I_t \wedge r.l \in L_{i}\right\}  \right|: L_{i} \in L \right\},
\end{equation}
where $\left| \cdot \right|$ denotes the cardinality of the set.
We define graph $\mathcal{G}^{(t)}=(\mathcal{V}^{(t)},\mathcal{E})$, where $\mathcal{V}^{(t)}$ is the set of node features in $I_{t}$ and $\mathcal{E}$ is the set of edges between nodes.
In here, each node is corresponded to a region in $L$.
If $L_i$ and $L_j$ are adjacent, $e_{ij} \in \mathcal{E}$ is defined as 1, otherwise 0.
A node features of $L_i$ in $I_t$ is defined by $\mathbf{v}^{(t)}_{i} \in \mathcal{V}^{(t)}$,
\begin{equation} \label{eq_v_def}
\mathbf{v}^{(t)}_{i} = [x^{(t)}_i, x^{(t-1)}_i, ..., x^{(t-T+1)}_i]^\top.
\end{equation}

\subsection{Demand Forecasting Problem}
A short-term demand forecasting model $\mathcal{F}$ predicts demand volumes in time interval $(t+1)$, using demand volumes until $(t)$.
We consider graph of demand data as input, thus the forecasting model with lag $T$ is
\begin{equation} \label{eq_model}
\hat { \textbf{x} } ^{ (t+1) } = \mathcal{F}(\textbf{x}^{(t)}, \textbf{x}^{(t-1)}, ..., \textbf{x}^{(t-T+1)}) = \mathcal{F}(\mathcal{V}^{(t)}, \mathcal{E}),
\end{equation}
where $\hat { \textbf{x} } ^{ (t+1) }$ is predicted demand volumes.
The predictive model is estimated to maximized likelihood of training data,
\begin{equation}
    p(X_{T+1}|X_{T}, X_{T-1}, ..., X_{1}),
\end{equation}
where $X_{i}$ denote i-th random variable in ordered sequence, $(X_{T}, X_{T-1}, ..., X_{1})$.
Note that the training of forecasting model $\mathcal{F}$ is not dependent on specific time stamp $t$, but optimized to predict next value from ordered sequence with fixed length $T$.

In general, there are an important remarks in time series modeling.
Forecasting models assume the training samples are from a stationary process, whose probability distribution is not changed over time $t$.
If time series have different probability distribution over time, some preprocessings, such as differencing or scaling, are needed to input sequences satisfy stationary condition \cite{wei2006time}.
When we use deep neural networks to model non-stationary time series, we expect neural networks can extract hidden \textit{stationary} features over time (see details in supplementary).

\section{Temporal-Guided Networks} \label{3}
\subsection{Graph Networks for Spatial Features with Permutational Invariance}
Convolutional layers learn to extract spatial features of a region from its adjacent regions, but a convolution is permutation-variant operation in general.
Then, it is dependent on the locational permutations and orderings of its neighborhood.

We claim that permutation-invariant operation is more efficient way to extract spatial correlations between a region and its neighborhood than convolution.
When we define spatial features of a region, the characteristics of its neighborhood and the proximity of them are enough to be considered, instead of their directionality.
For example, the proximity of a subway station from a region is enough to define features of the region than where the station is in west or east of the target region.
Convolution considers locational permutations of neighborhood and requires different filters by permutations of neighborhood in general.
It can increase the number of trainable parameters unnecessarily and result in overfitting when the training data are limited.

We use permutation-invariant operation to aggregate features of adjacent regions to each region.
When a spatial feature of each region is extracted, the directionality of its neighborhood does not considered in permutation-invariant operation.
It can efficiently reduce the number of trainable parameters and prevent from overfitting problem, maintaining or improving forecasting results on test data.
For simplicity of notation, we use $\mathbf{v}_i$ instead of Equation~\ref{eq_v_def}.
\begin{equation}
    \mathbf{h}^k_{\mathcal{N}(i)} = \textrm{MEAN}(\left\{{\mathbf{W}^{k}_{\mathcal{N}}} \cdot \mathbf{v}^{k-1}_u, \forall u \in \mathcal{N}(i)\right\}) \label{eq_nb_h}
\end{equation}
\begin{equation}
    \mathbf{h}^k_{i} = \mathbf{W}^{k}_{v} \cdot \mathbf{v}^{k-1}_i \label{eq_v_h},
\end{equation}
where $\mathbf{v}^{k-1}_i$ is feature vector of node $i$ in (k-1)th layer,$\mathcal{N}(i)$ is the neighborhood regions of region $i$, trainable parameters in k-th layer, $\mathbf{W}^k$.
Note that Equation~\ref{eq_nb_h} receive messages from feature vectors of neighbor regions and use permutation-invariant operation to aggregate them.
Feature vector of node $i$ is calculated by a fully connected layer, combining aggregation of its neighborhood and linear transformation of the node.
\begin{equation} \label{eq_concat}
    \mathbf{v}^{k}_i = \textrm{ReLU}(\mathbf{W}^{k} \cdot [\mathbf{v}^{k-1}_i, \mathbf{h}^{k}_{\mathcal{N}(i)}+\mathbf{h}^{k}_{i}]),
\end{equation}
where $[\cdot, \cdot]$ and $+$ are concatenation and element-wise summation. The concatenation in Equation~\ref{eq_concat} is a kind of skip connection and helps model learn with feature reuse and alleviation of gradient vanishing problem \cite{huang2017densely}.
All trainable parameters in each layer are shared over every node.

After $K$ layers of graph networks, demand volume of region $i$ at time $(t+1)$ is predicted as
\begin{equation} \label{eq_output}
    x^{(t+1)}_i = \textrm{ReLU}(\mathbf{w}^\top \textrm{ReLU}(\mathbf{W}^{K+1}\cdot[\mathbf{v}^{K}_{i}, \mathbf{q}_{i}])),
\end{equation}
where $\mathbf{q}_{i}$ is feature vector of region $i$ from external data sources and is explained in next section.
ReLU is also used in output layer to produce positive demand values.
Note that above operations can be generalized to different tessellations of city, such as hexagonal \cite{ke2018hexagon} or irregular patterns \cite{davis2018taxi}.

\subsection{Temporal-Guided Embedding}

Temporal-guided embedding is proposed to learn temporal contexts in training data explicitly and consider temporally recurrent patterns. 
We assume that the combination of immediate past data and learned temporal context can substitute for days/weeks ago histories to capture temporally recurrent patterns.
The temporal-guided embedding at time $t+1$ is defined by
\begin{equation}
TGE_{(t+1)} = f_{TGE}(\mathbf{\tau}_{t+1})    
\end{equation}
where $\tau_{t+1}$ is a $0/1$ categorical variable, which can represent temporal information of time $t+1$.
For example, we can use the concatenation of four one-hot vectors, which are time-of-day, day-of-week, holiday, and the day before holiday information of time $t+1$, to represent temporal information.
Fully connected layer, $f_{\textrm{TGE}}$, outputs learn to predict distributed representation of temporal information of $t+1$ in training time.

The temporal-guided embedding is concatenated into the input of model 
\begin{equation}
\hat{\mathbf{x}} ^{(t+1)} = \mathcal{F}(\mathcal{V}^{(t)}_{TGE}, \mathcal{E}), \\
\end{equation}
\begin{equation}
\mathcal{V}^{(t)}_{TGE} = \left\{ [\mathbf{v}^{(t)}_{i}, TGE_{(t+1)}]: \forall L_{i} \in L \right\},
\end{equation}
where $[\cdot, \cdot]$ is feature-wise concatenation and temporal information of $t+1$ is available at time t. 

We assume that temporal-guided embedding make the model learn conditional distribution on temporal contexts of forecasting target.
The, TGNet can extract hidden \textit{stationary} features from nonstationary demand patterns, conditioned on the learned embedding of temporal contexts.
Learning conditional distribution of images on labels \cite{mirza2014conditional} or words on positions \cite{vaswani2017attention}, exists and shows significant improvement of performance.
To the best of our knowledge, temporal-guided embedding is simple idea, but is the first approach in time series domain to learn temporal context explicitly and show interpretable visualization of temporal contexts in training data.
Note that determining the periodicity and seasonality in long-term histories is a heuristic or hand-craft procedure by partial ACF, but temporal-guided embedding can learn temporal contexts directly and replace the procedures.

\subsection{Late Fusion with External Data Sources}
Orthogonal to capturing of complex spatiotemporal patterns in demand data, the forecasting results can be improved by incorporating external data such as meteorological, traffic flow, or event information \cite{tong2017simpler,zhang2016dnn,zhang2017deep,yao2018deep,yao2018modeling}.
In this paper, we do not use external data sources to improve our results. 
However, we explain how our model architecture incorporates data on other domains.

As an example, drop-off volumes in past are used to improve demand forecasting results, because drop-off in a region might be changed into demands in future \cite{yao2018modeling,vahedian2019predicting}.
Feature vectors of drop-off patterns $\mathbf{q}_i$ are extracted by graph networks in the same manner and concatenated into the features from demand (Equation~\ref{eq_output}).
This type of late fusion is a common approach to combine heterogeneous data sources from multimodality \cite{zadeh2017tensor,anderson2018bottom,ku2018joint}. 
We expect that various external data can be incorporated by this manner to improve the results in future work.

\section{Experiment} \label{4}
\subsection{Experimental Setting}
\textbf{Datasets   }
Three real-world datasets (NYC-bike, NYC-taxi, and SEO-taxi) are used for evaluation.
The first two datasets are open publicly and SEO-taxi is private \cite{nycb2017data,taxi2017tlc}.
40 days (4 months) data is used for training purpose, and the remaining 20 days (2 months) are tested in NYC (SEO) dataset.
The details of datasets are described in supplementary material.

\textbf{Evaluation   } 
We use two evaluation metrics to measure the accuracy of forecasting results: mean absolute percentage error (MAPE) and root mean squared error (RMSE).
We follow same evaluation method with \cite{yao2018modeling,yao2018deep} for fair comparison and excluded samples with less value than $k$.
It is known as common practice in industry and academia, because real-world applications have little interest in such low-volume samples.
In all tables in this paper, the mean performances with ten repeats are reported and bold means statistical significance.

\textbf{Implementations   } 
NYC and Seoul are divided into 10$\times$20 and 50$\times$50 regions respectively, considering the area size of cities.
Each region is about 700 m$\times$700 m.
Time intervals are divided into 30 minutes.
Demands and drop-off volumes in previous 8 and 16 time intervals (4 and 8 hours) are used to forecast demands in the next time interval (30 minutes).

Batch normalization \cite{ioffe2015batch} and dropout \cite{srivastava2014dropout} with $p=0.1$ are used in every layers.
TGNet has 7 hidden layers and the number of hidden neuron of the first hidden layer is 32 (NYC) or 64 (SEO).
We doubly increase the number of neurons in each layer.
We attach the source codes with Tensorflow 1.17.0 \cite{abadi2016tensorflow} and Keras 2.22.2 \cite{chollet2015keras}, and detail explanations in supplementary, including the number of layers and hidden neurons.

\textbf{Training   } 
We use two types of loss to train TGNet.
We used L2 loss (mean square error) first and change the loss to L1 (mean absolute error).
L1 loss is more robust to the anomalies in the real time series \cite{lai2018modeling}, but the optimization process was not stable experimentally.
Initial training with L2 loss makes the optimization with L1 loss stable.
TGNet is trained with Adam optimizer \cite{kingma2014adam} using 0.01 learning and decay rate.
20 \% of training data are used for validation and early-stopping is applied to select an optimal model.
That is $1,523/381/477$ ($4,595/1,149/2,912$) numbers of samples in NYC (SEO) are used for training/valid/test.
Two Tesla P40 GPUs are used and about 2 (26) hours are takes for training NYC (SEO) dataset.

\textbf{Compared Methods   } 
We compare TGNet with statistical and state-of-the-art deep learning methods for spatiotemporal data:
ARIMA, XGBoost \cite{chen2016xgboost}, STResNet \cite{zhang2017deep}, DMVST-Net \cite{yao2018deep}, and STDN \cite{yao2018modeling}.

\begin{table*}[]
\small
\caption {Comparison of Forecasting Results on NYC-bike, NYC-taxi, and SEO-taxi. Average values with ten repeats are reported in the table and bold means statistical significance to STDN \cite{yao2018modeling}.}
\label{tab:overall_result}
\centering
\begin{tabular}{l|ccccccccc}
\toprule
\multirow{2}{*}{Method} & \multicolumn{2}{c}{NYC-bike}            & \multicolumn{2}{c}{NYC-taxi}              & \multicolumn{2}{c}{SEO-taxi} & \multicolumn{2}{c}{(NYC)} & External \\
                        & RMSE               & MAPE(\%)            & RMSE                & MAPE(\%)            & RMSE          & MAPE(\%)   & \multicolumn{2}{c}{Params} & Data     \\ \hline
ARIMA                   & 11.53              & 27.82               & 36.53               & 28.51               &    48.92      &     56.43  & \multicolumn{2}{c}{-}  & X  \\
XGBoost                 & 9.57               & 23.52               & 26.07               & 19.35               &    32.09       &   45.75 & \multicolumn{2}{c}{-} & X\\
STResNet                & 9.80           & 25.06           & 26.23           & 21.13           & -             & -  & \multicolumn{2}{c}{4.8 M} & O            \\
DMVST-Net              & 9.14           & 22.20           & 25.74           & 17.38           & -             & -  & \multicolumn{2}{c}{1.5 M} & O         \\
STDN                & 8.85           & 21.84          & 24.10           & 16.30           & -             & -   & \multicolumn{2}{c}{9.4 M} & O           \\ \hline
GN                   & 9.09  & 22.51  & 23.75  & 15.43  & 28.10 & 37.31 & \multicolumn{2}{c}{0.41 M} & X\\ 
GN+TGE                   & 8.88 & 22.37  & 22.81 & 14.99 & 25.96 & \textbf{35.67} & \multicolumn{2}{c}{0.42 M} & X\\  
TGNet                   & 8.84 & 21.92 & \textbf{22.75} & \textbf{14.83} & \textbf{25.35} & 35.72 & \multicolumn{2}{c}{0.48 M} & X\\ 
\bottomrule
\end{tabular}
\end{table*}

\subsection{Forecasting Performances Analysis}
We report the forecasting performances of TGNet and other compared models with ten repeats on NYC-bike, NYC-taxi, and SEO-taxi datasets in Table~\ref{tab:overall_result}.
In evaluation, the samples with demand volume less than 11 are eliminated.
The traditional time series model, ARIMA, shows the lowest accuracy on all datasets, because it cannot consider spatial correlations and complex non-linearity in demand patterns.
XGBoost shows better performances than statistical time series model.

Recent deep learning models outperform ARIMA and XGboost, capturing complex spatiotemporal correlations in the datasets.
The most recent model, STDN, shows the best performances on NYC-bike and NYC-taxi datasets among baseline methods.
STDN outperforms other baseline methods, because STDN utilizes long-term histories of demands and various modules such as periodic and seasonal inputs, local CNN and LSTM with periodically shifted attention, and external data (weather and traffic in/out flow).

\textbf{Efficiency and Accuracy  } TGNet has about 20 times smaller number of trainable parameters (475,543) than STDN (9,446,274).
However, TGNet shows better performances than all compared models, which use complex model architectures of large number of parameters and external data sources.
TGNet significantly reduces the number of trainable parameters, extracting spatiotemporal features by graph networks and temporal-guided embedding.
When we consider that TGNet is a simple baseline model, the competitive results are notable.

\textbf{Use of External Data sources  } Also, we note that the results of TGNet are achieved without external data sources, such as meteorological, traffic flow, or event information, which can improve overall forecasting results.
The performances on NYC-bike are not much better than those of STDN, because the demand patterns of bike are highly dependent on meteorological situations.
We expect that our forecasting performances can be more improved in future by external data sources.

\textbf{On Large-scale Dataset   } SEO-taxi dataset has 12.5 times larger regions and 3 times longer period than NYC datasets.
TGNet can learn SEO-taxi dataset only by increasing the number of hidden neurons in each layer, but, to the best our effort, we fail to train the compared models with deep learning (STResNet, DMVST-Net, and STDN) on SEO-taxi dataset.
We accept the failure training of compared models, because they were not validated on large-scale datasets in their own works \cite{zhang2017deep,yao2018modeling,yao2018deep} and the numbers of trainable parameters are excessively large as NYC datasets' sample number (1,523) and shape ($10\times 20$).
Simple and efficient model architecture is compelling to generalize from the scale of datasets.

In summary, TGNet has more efficient and competitive results as a new baseline model for demand forecasting and we expect that our model will be improved with combining external data sources and adding various modules.

\begin{table}[]
\small
\caption {Comparison of TGNets with Convolutional and Graph Networks.}
\label{tab:ablation_cnn}
\centering
\begin{tabular}{lc|ccc}
\toprule
                                                                       &       & TGNet-C-A & TGNet-C-B & TGNet   \\ \hline
\multirow{2}{*}{\begin{tabular}[c]{@{}c@{}}NYC-\\bike\end{tabular}}   & RMSE  & 9.17      & 9.10      & \textbf{8.84}    \\
                                                                       & MAPE  & 22.19     & 22.13     & \textbf{21.92}   \\ \hline
\multirow{2}{*}{\begin{tabular}[c]{@{}c@{}}NYC-\\taxi\end{tabular}}   & RMSE  & 23.68     & 23.03     & \textbf{22.75}   \\
                                                                       & MAPE  & 15.01     & 15.15     & \textbf{14.83}   \\ \hline
\multicolumn{2}{c|}{\# of parameters} & 737,751   & 876,951   & 475,543 \\
\bottomrule
\end{tabular}
\end{table}


\subsection{Ablation Study}
\subsubsection{Effects of Graph Networks}
We compared two types of variants of TGNet to know the effect of permutational-invariant operation:

\textbf{TGNet-C-A} uses convolutional neural networks instead of graph networks in TGNet.

\textbf{TGNet-C-B} substitutes aggregation operation (Equation~\ref{eq_nb_h}) with 3$\times$3 convolution operation and keeps other operations in graph networks same.

Convolution layer can also learn permutation-invariant operation, but the results in Table~\ref{tab:ablation_cnn} show that permutation-invariant operation can be enough to learn spatial relationship between adjacent regions.
The forecasting performances of graph networks are also better than the other convolutional variants, and the number of parameters are about 1.5-2 times smaller than the others.
Note that the performances of convolutional variants (TGNet-C-A and TGNet-C-B) are also competitive with DMVST-Net and STDN.
The results imply that extracting spatiotemporal features at once is more effective in demand forecasting than extracting spatial and temporal features separately \cite{zhang2017deep,yao2018deep,yao2018modeling}.

We conclude that extracting complex spatiotemporal features with permutation-invariant graph network can be efficient and effective way to model spatiotemporal demands.

\subsubsection{Effects of Temporal-Guided Embedding}
Temporal-guided embedding improves forecasting performances (both RMSE and MAPE) on all real-world datasets in Table~\ref{tab:overall_result}.
The performance gains are significant when we consider the difference of performances and the number of parameters between other models.
For example, in the case of NYC-taxi (bike), temporal-guided embedding reduces MAPE by 0.94 (0.21), but the number of trainable parameters only increase by 9,000 to learn temporal-guided embedding.
Meanwhile, STDN reduces MAPE by 1.64 (0.29) than DMVST-Net and DMVST-Net reduces MAPE by 0.49 (0.66) than STResNet.
Temporal-guided embedding is simple idea to encode temporal information, which help TGNet extracts stationary features from non-stationary time series according to temporal contexts of input.
In spite of the simplicity, the effectiveness of temporal-guided embedding is significant to improve overall forecasting results.

\subsection{Visualization of Temporal-Guided Embedding}
If the performance gains of temporal-guided embedding are from understanding of temporal contexts of input data, we expect that the embeddings show meaningful visualization, which have interpretable results of temporal contexts in training data.
Note that the inputs of temporal-guided embedding is $0/1$ categorical variable and they are not mutually correlated to each other.
For example, the time-of-day vectors of 5 a.m. and 6 a.m. are independent, because the input is one-hot encoded.

We find three remarkable results that temporal-guided embedding actually learns and extracts temporal contexts from input data.
Firstly, the embeddings of adjacent time-of-day are located adjacent to each other in embedding space (see the results in supplementary material).
High correlation between adjacent time is basic assumption of autoregressive models (including LSTMs) and temporal-guided embedding can learn the basic idea of time-series modeling \cite{wei2006time}.

\begin{figure}[h]
\centering
\includegraphics[width=8.5cm]{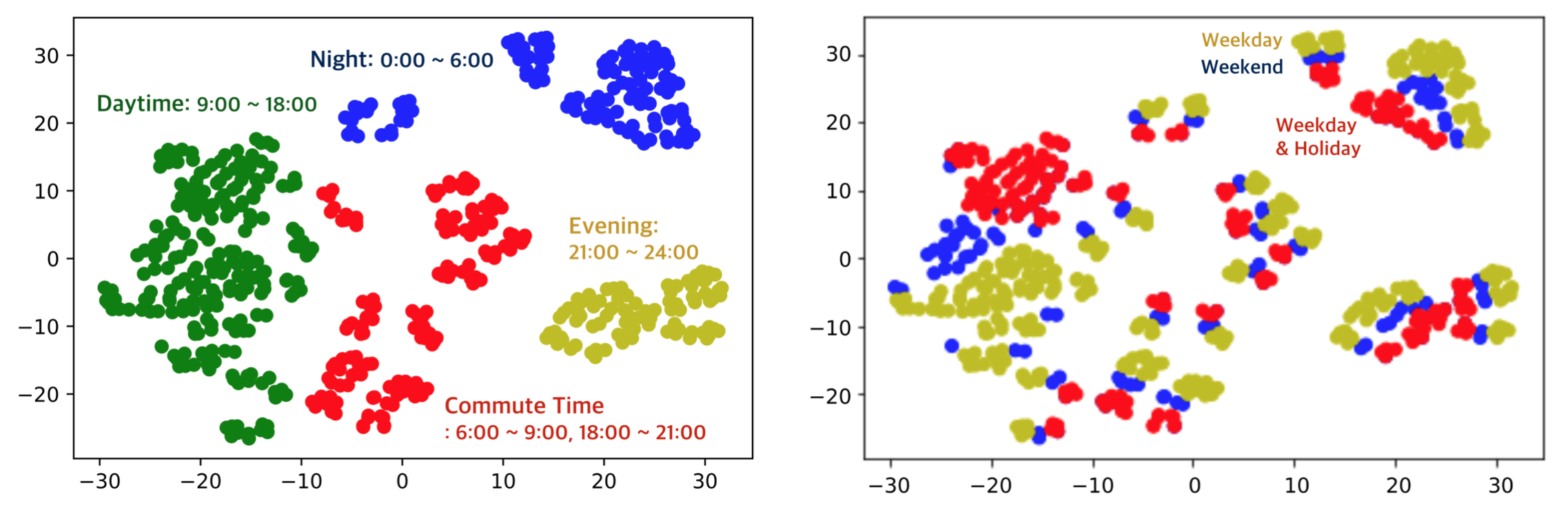}
\caption{t-SNE visualization of temporal-guided embedding from SEO-taxi dataset. The embedding vectors of time-of-day are divided into 4 clusters: commute time, daytime, evening, and night (left). Weekday and weekend vector are divided regardless of time-of-day, and the holiday (but weekday) vectors are adjacent to weekend vectors (right).}
\label{tge_visualization3}
\end{figure}

Secondly, the clusters of embedding vectors on time-of-day represent different temporal contexts of demand patterns in a day.
The embeddings of time-of-day vector are classified into four clusters: commute time, daytime, evening, and night (Figure~\ref{tge_visualization3} left).
The clustering result is analogous to the way that people understand daily taxi demand patterns based on their common lifestyle.

Lastly, temporal-guided embedding learns the concept of day-of-week and holiday.
The locations of weekday and weekend vectors are strictly divided. 
If a day-of-week is weekday and it is holiday, the embedding is adjacent to weekend vector, because holiday and weekend demand patterns are similar (Figure~\ref{tge_visualization3} right).
In summary, temporal-guided embedding not only improves forecasting results, but also can learn temporally contextual meaning of demand patterns from training dataset and show interpretable visualizations.

\begin{table*}[t]
\small
\caption {Performances on atypical samples in NYC- and SEO-taxi datasets}
\label{tab:atypical_result}
\centering
\begin{tabular}{l|cccc|cccc}
\toprule
                        & \multicolumn{4}{c}{NYC-taxi} & \multicolumn{4}{|c}{SEO-taxi}\\ \hline
\multirow{2}{*}{Method} & \multicolumn{2}{c}{RMSE}                         & \multicolumn{2}{c}{MAPE (\%)} & \multicolumn{2}{|c}{RMSE}                         & \multicolumn{2}{c}{MAPE (\%)}                    \\
                        & top 1 \%       & top 5 \%      & top 1 \%       & top 5 \%       & top 1 \%       & top 5 \%      & top 1 \%       & top 5 \% \\ \hline
STResNet                & 224.50         & 217.72          & 154.06         & 157.72         & N/A & N/A & N/A & N/A\\  
STDN                    & 210.34         & 203.11            & 90.55          & 89.71        & N/A & N/A & N/A & N/A\\ \hline
GN                     & 21.15          & 20.36                & 28.75          & 29.62          & 39.91          & 30.99          & 47.32          & 48.08          \\
GN + TGE               & 20.79          & 20.03                & 27.51          & 28.36               & 37.18          & 28.96          & \textbf{45.86}          & \textbf{46.78} \\
TGNet                   & \textbf{19.64} & \textbf{18.83}  & \textbf{27.43} & \textbf{28.23} & \textbf{36.37} & \textbf{28.19} & 46.16 & 47.16\\
\bottomrule
\end{tabular}
\end{table*}

\subsection{Forecasting when Atypical Events Occurs} 
In practice, short-term demand forecasting is important when atypical events, which have abnormally large demands than usual, occur.
For example, bad performances on these situations cause fatal problem of supply-demand mismatch and ride-hailing services can be connected to service failure.
Abnormally high values are non-repetitive and have different patterns from majority of samples \cite{vahedian2019predicting}.
Thus, they are hard to be learned by forecasting model, because they do not appear often in training time in spite of their importance in real-world.

We select atypical event samples in test data, which are larger than each threshold in each region.
We also set different thresholds according to time-of-day, weekend, and holiday information by each region to consider different spatial and temporal contexts.
Samples above top 1 \% and 5 \% thresholds in each region are selected.
All thresholds are larger than 10 times of the standard deviation from mean demand of each region.
We also identify that atypical event samples, such as concert, festival, or academic conferences, are included.

Recent deep learning models (STResNet and STDN) of demand forecasting cannot learn minority samples with extremely large values (Table~\ref{tab:atypical_result}).
Excessive number of trainable parameters results in overfitting, and the models are not generalized to minority samples, which rarely appear during training time.
However, the forecasting results of TGNet are also well generalized to atypical event samples, and have acceptable performances.
We also find that drop-off volumes can be helpful in atypical event situation to improve forecasting results, because past surge of drop-off volumes can be converted to a future demand \cite{vahedian2019predicting}.
To the best our knowledge, it is the first attempt to evaluate a model on atypical event samples.
We claim that forecasting models need to be evaluated not only with average performances, but also on unusual event samples for real-world applications.

\section{Conclusions} \label{6}
In this study, we propose an efficient baseline model of short-term demand forecasting, temporal-guided network (TGNet), which consists of graph neural networks and temporal-guided embedding.
Our model is evaluated on three real-world datasets and achieves competitive performances with 20 times smaller number of trainable parameters than a recent state-of-the-art model \cite{yao2018modeling}.
As a baseline model, the performances of our model are achieved without complex architecture, input of long-term histories, and external data sources.
We show that spatial features of each region are enough to be invariant to locational permutation of its adjacent regions.
Temporal-guided embedding can learn to encode temporal contexts in training data and improve overall forecasting performances, showing interpretable visualization of temporal contexts.
We also show that previous approaches have a problem of generalization on atypical events samples, which have extremely large values and do not appear frequently in training time, but are important in practice.
However, our model have more robust performances on atypical event samples than other compared models.

In future work, TGNet can be improved by utilizing external data sources, long-term histories, and various architecture modification, as the previous models with the combination of CNNs and LSTMs are advanced.
We expect that our model can be a new backbone of building block for spatiotemporal demand forecasting models.

\bibliographystyle{plain}
\bibliography{main}

\clearpage
\appendix
\section{Implementation}
\subsection{Implementation Details}
Our codes are based on Tensorflow 1.7.0 \cite{abadi2016tensorflow} and we used high-level API, Keras 2.2.2 \cite{chollet2015keras}.
The source codes including README.txt are available on supplementary material.
There are six hidden layers before fully-connected layer (equation (7)) and two layers are used for taxi drop-off volumes.
We use a 2d average pooling layer with 2x2 kernel after GN 1 layer for computational efficiency.
Skip connections like \cite{ronneberger2015u} are used to alleviate gradient vanishing problem \cite{huang2017densely}.

\begin{figure}[h]
\centering
\includegraphics[width=13.0cm]{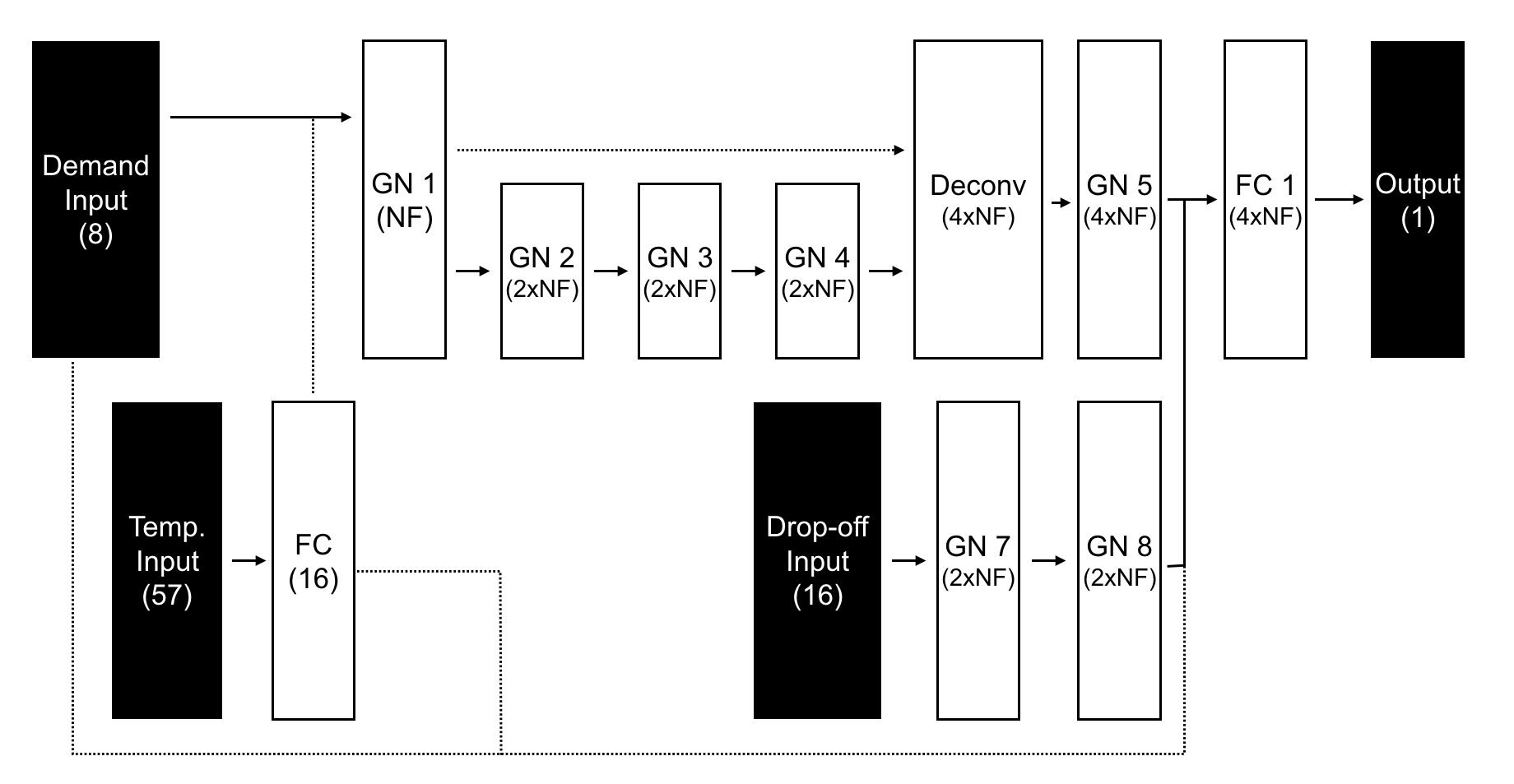}
\caption{Overall model architecture of TGNet. Dot line means skip connection by concatenation and NF means number of states in each layer.}
\label{model_arch}
\end{figure}

The number of hidden neurons of first layer (NF in Figure~\ref{model_arch}) is 32 in NYC datasets.
We increase the number of neurons twice because SEO-taxi dataset is relatively larger scale than NYC datasets.
Batch Normalization and dropout are used in each layer.

\subsection{Methods for Comparison}
We compare the performances of TGNet with existing demand forecasting models from spatiotemporal data and describe them in this section.
We follow the hyperparameters in original papers, but adjust learning rates for training.

\textbf{ARIMA}: Autoregressive Integrated Moving Average (ARIMA) is traditional model for non-stationary time series.
We use auto ARIMA function in R\cite{team2013r} to fit each dataset.

\textbf{XGBoost} \cite{chen2016xgboost}: XGBoost is a popular tool to train a boosted tree. 
The number of trees is 500, max depth is 4, and subsample rate is 0.6.

\textbf{ST-ResNet} \cite{zhang2017deep}: ST-ResNet is a CNN-based model with residual blocks \cite{he2016deep}.
They uses various past time step as temporal closeness, periodic, and seasonal inputs to capture temporally recurrent patterns.
ResNet is used to extract hidden representations of each input (a image at each time step) and they concatenate all feature maps before prediction of future demand.

\textbf{DMVST-Net} \cite{yao2018deep}: DMVST-Net models spatial, temporal, and semantic view through local CNN, LSTM, and graph embedding.
They do not forecast demands of all target regions at once, but predict future demand of each region independently.
After convolutional layers extract spatial feature of input image at each time, the feature maps are entered into LSTM layers to extract temporal features.

\textbf{STDN} \cite{yao2018modeling}: STDN is based on DMVST-Net \cite{yao2018deep}, and add some parts to improve forecasting results.
Temporal closeness, periodic, and seasonal inputs are used to model temporally recurrent pattern and periodically shifted attention is proposed to deal with long sequence. 
Traffic flow and taxi drop-off volumes are also used with flow gating mechanisms.

\subsection{Hyperparameter Search}
We used greed search with various setting below and determine optimal hyperparameters.
The bold means selected ones.
The number of hidden neurons in the first layer, in SEO-taxi, was 64 and the others were same.

\textbf{Learning Rate}: Learning rate for Adam optimizer. $\left\{0.1, \textbf{0.01}, 0.001 \right\}$

\textbf{Decaying Rate}: Decaying rate for Adam optimizer. $\left\{0.1, \textbf{0.01}, 0.001 \right\}$

\textbf{Number of Hidden Neurons}: The number of convolutional filters in first hidden layer.
The Filter numbers in other layers have same ratio with optimal one, mentioned above.
$\left\{16, \textbf{32}, 64, 128 \right\}$

\textbf{Number of Hidden Neurons for Drop-off}: The number of convolutional filters for encoding of drop-off volumes. 
$\left\{16, 32, \textbf{64}, 128 \right\}$

\textbf{Batch Size}: mini-batch size for model update.
$\left\{16, 32, 64, \textbf{128} \right\}$

\textbf{Dimensionality of Temporal-Guided Embedding}: the number of dimension for temporal-guided embedding.
$\left\{2, 4, 8, \textbf{16}, 32 \right\}$

\section{Evaluation}
We introduce three real-world datasets, which are used to evaluate our model, and other evaluation details.
The region of NYC is divided into 10$\times$20 grid and the region of Seoul is into 50$\times$50 grid.
A grid cell covers about 700 m$\times$700 m.
A time interval is 30 minutes in this study.

\subsection{Dataset Description}
\textbf{NYC-bike} NYC-bike dataset contains the number of rents and returns of bike in NYC from 07/01/2016 to 08/29/2016.
The first 40 days are used for training purpose and the remaining 20 days are as test.
This dataset is not about taxi demand, but we also evaluate this dataset to generalize our model as spatiotemporal demand forecasting model.
The demand patterns on bike are vulnerable to weather condition.
For example, if a day is rainy, there is no demand of bike.
In this paper, we do not use external data, including weather, but we show our model have competitive performances on other baselines with external data.

\textbf{NYC-taxi} NYC-Taxi dataset contains taxi pick-up and drop-off records of NYC in from 01/01/2015 to 03/01/2015.
The first 40 days data is used for training purpose, and the remaining 20 days are tested. 

\textbf{SEO-taxi} SEO-taxi dataset contains ride request and drop-off records in Seoul, South Korea.
This data are provided from a on-demand ride-hailing service provider and is private.
The period of dataset is from 01/01/2018 to 06/30/2018 and the first 4 months data are used for training and the remaining for test.
SEO-taxi dataset is relatively large-scale, because the area of Seoul (50$\times$50 grid) is larger than NYC (10$\times$20) and the period is also longer than NYC-bike and NYC-taxi.
We found that other baselines could not learn SEO-taxi in hyperparameter settings described above to the our best effort.

\subsection{Dataset Details}
In this paper, the other deep learning models can't learn SEO-taxi dataset because it is large-scale and more sparse and complex.
The dataset is private now, so we attach the comparison of three datasets, in statistics.
We will upload SEO-taxi dataset, if it is free to the security issue.

In Table~\ref{tab:stat_db} shows the statistics of datasets: NYC-bike, NYC-taxi, and SEO-taxi.
We set a time interval as 30 minutes.
The period of NYC datasets is 60 days and 2,880 time steps with 10$\times$20 regions.
On the other hand, SEO-taxi dataset is 181 days and 8,688 steps with 50$\times$50 regions.
SEO-taxi has lower mean and standard deviation than NYC-taxi, but the maximum demand volume of SEO-taxi is much larger.
We consider that SEO-taxi has more sparse and complex dynamics of taxi demands with large-scale of regions and times.

\begin{table}[h]
\caption{Statistics of Datasets. Mean value of SEO-taxi is not provided because of security issue.}
\label{tab:stat_db}
\centering
\begin{tabular}{c|ccccccc}
\toprule
Dataset  & Period                                                              & Regions & Mean & Median & Std    & Min & Max  \\ \hline
NYC-bike & \begin{tabular}[c]{@{}c@{}}07/01/2016-\\ 08/29/2016\end{tabular} & 10 x 20 & 4.52 & 0      & 14.33  & 0   & 307  \\ \hline
NYC-taxi & \begin{tabular}[c]{@{}c@{}}01/01/2015-\\ 03/01/2015\end{tabular} & 10 x 20 & 38.8 & 0      & 107.71 & 0   & 1,149 \\ \hline
SEO-taxi & \begin{tabular}[c]{@{}c@{}}01/01/2018-\\ 06/30/2018\end{tabular} & 50 x 50 & -    & 0      & 18.27  & 0   & 4,491 \\
\bottomrule
\end{tabular}
\end{table}

\subsection{Performance Measures}
Two evaluation metrics measure the performances of forecasting models: Mean absolute percentage error (MAPE) and root mean squared error (RMSE). 
In evaluation, the samples with value less than $k$ are excluded as a common practice in industry and academia \cite{yao2018deep,yao2018modeling}, because they are of little interest in real-world applications.
Let $\underline{X}^{(t)}=\left\{ {X}^{(t)}_{ij} | X^{(t)}_{ij} \ge k \right\}$ be the set of filtered samples, then the performance measures are given by
\begin{gather}
\textrm{RMSE} = \frac{1}{|{\underline {X}}^{(t)}_{ij}|}\sum_t{\sum_{i,j} (\hat{\underline {X}}^{(t)}_{ij}-\underline{X}^{(t)}_{ij})^2}, \\
\textrm{MAPE} = \frac{1}{|\underline {X}^{(t)}_{ij}|} \sum_t\sum_{i,j} \left|\frac{\hat{\underline {X}}^{(t)}_{ij} - \underline {X}^{(t)}_{ij}}{\underline {X}^{(t)}_{ij}}\right|.
\end{gather}
MAPE and RMSE tend to be sensitive to low and large value samples, respectively.
For extreme cases with one sample, if model prediction is 3 when the ground-truth is 1, MAPE is 200 \% and RMSE is 2.
On the other hand, if model prediction is 500 when the ground-truth is 1,000, MAPE is 50 \% and RMSE is 500.
Because of these characteristics, both measures are compared together.

\section{Temporal-Guided Embedding}
\subsection{Input of Temporal-Guided Embedding}
The input of temporal-guided embedding is concatenation of four one-hot vectors (time-of-day, day-of-week, holiday or not, and the day before holiday or not) and the input vector is 0/1 categorical variables.
The detail explanations are in Table~\ref{tab:TGE}.
For example of time-of-day, one-hot vector means all time-of-days are independent to each other and there are no correlation between time-of-day vectors.
However, we expect that temporal-guided embedding can learn distributed representations of temporal contexts in the process of learning how to forecast taxi demand and understanding the characteristics of time series.
\begin{table}[h]
\caption{Categorical Inputs of Temporal-Guided Embedding: The input of temporal-guided embedding is 0/1 categorical vector that concatenates four one-hot vectors (time-of-day, day-of-week, holiday, and the day before holiday).}
\label{tab:TGE}
\centering
\begin{tabular}{l|c|c}
    \hline
    Type &  Dimensionality & Explanation\\
    \hline
    Time of Day   &   48 & 30 Minutes\\
    Day of Week    &   7 & MTWTFSS \\
    Holiday    &   1 & Holiday or not \\
    Bef. Holiday    &   1 & The day bef. holiday or not \\
    \hline
    Total    &   57 &  0/1 variables\\
    \hline
\end{tabular}
\end{table}

\subsection{Visualization of Temporal-Guided Embeddings}
We visualize learned temporal-guided embedding to investigate whether the embeddings are interpretable or not.
We assumed that temporal-guided embeddings can have meaningful insights or visualization over performance gains.
For visualization, we use t-SNE \cite{maaten2008visualizing} in scikit-learn 0.19.1.
Learning rate is 1,000 and other hyperparameters are set by default.

\begin{figure}
\centering
\includegraphics[width=13.0cm]{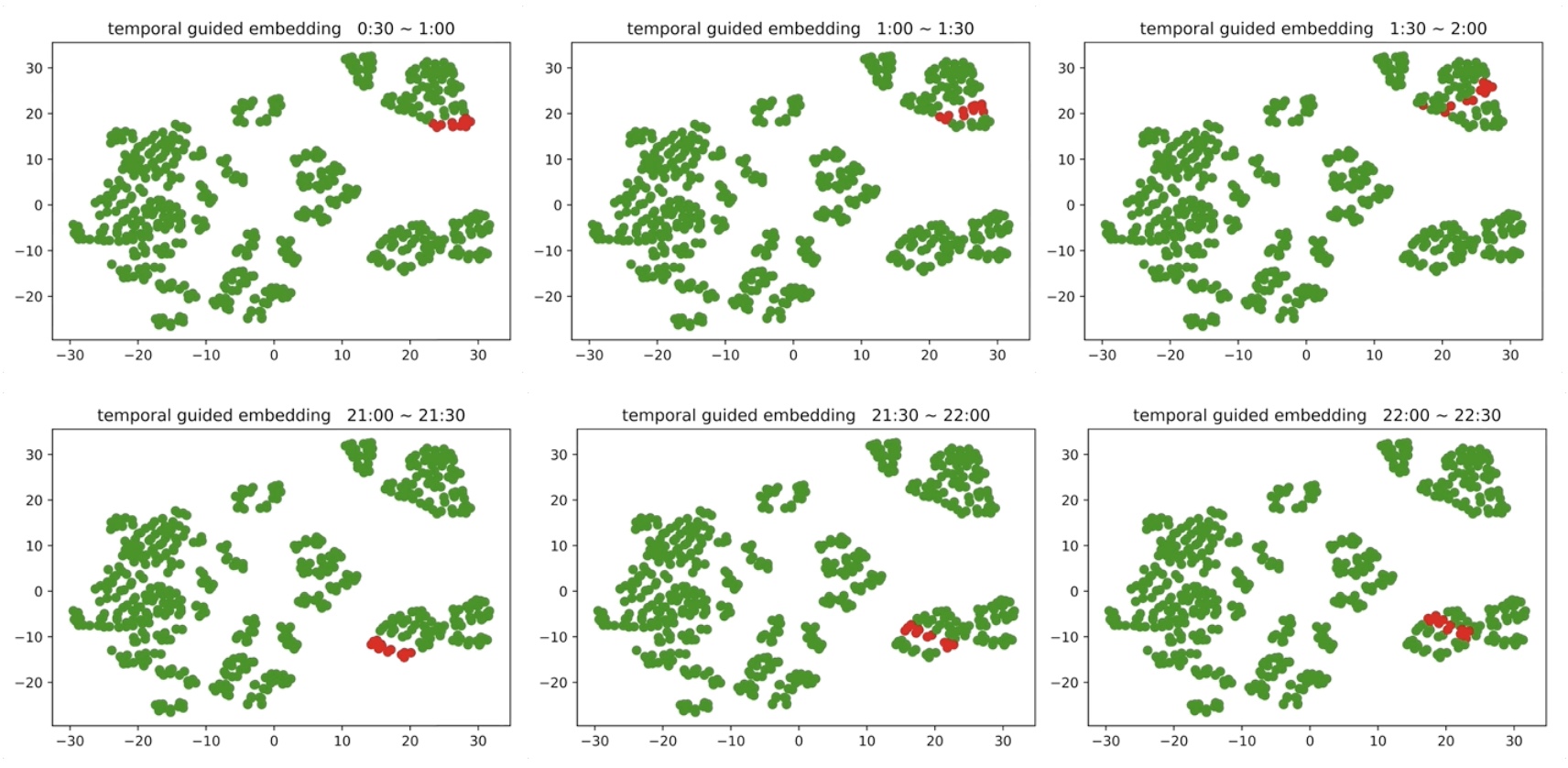}
\caption{Some example of temporal-guided embedding according to different time-of-days from SEO-taxi dataset. Adjacent time-of-day vectors are adjacent to each others.}
\label{tge_adj}
\end{figure}

We visualize some examples of temporal-guided embedding of different time-of-day vectors from SEO-taxi datasets in Figure~\ref{tge_adj}.
We find that temporal-guided embedding learn to locate the adjacent time-of-day vectors nearby each other.
The results are similar with human's understanding about the basic concept of time, because people naturally assume that events as adjacent time are strongly correlated with.
The assumption is also applied in sequential modeling with recurrent layers as relational inductive biases of series. 
The embeddings of remaining time intervals are available in supplementary material.

\begin{figure}[h]
\centering
\includegraphics[width=14cm, height=3cm]{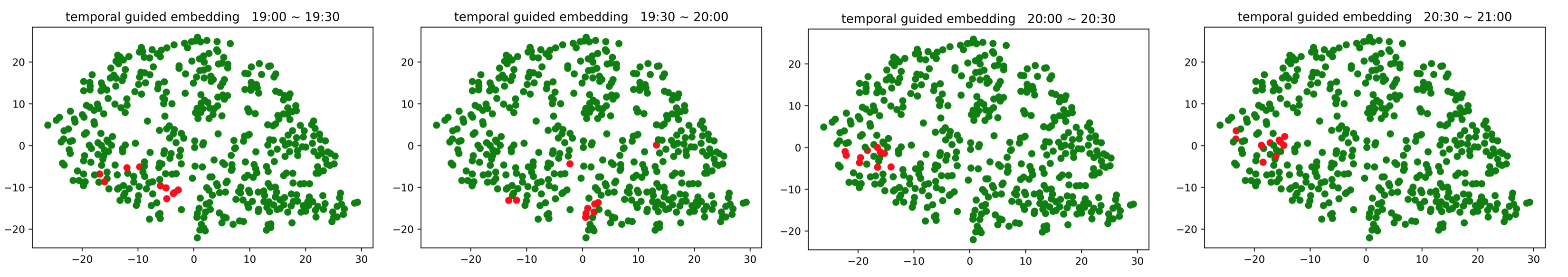}
\caption{t-SNE visualization of temporal-guided embedding of NYC-taxi. Adjacent time-of-day vectors are located nearby in the embedding space.}
\label{tge_nyc_tod}
\end{figure}

\begin{figure}[h]
\centering
\includegraphics[width=7cm]{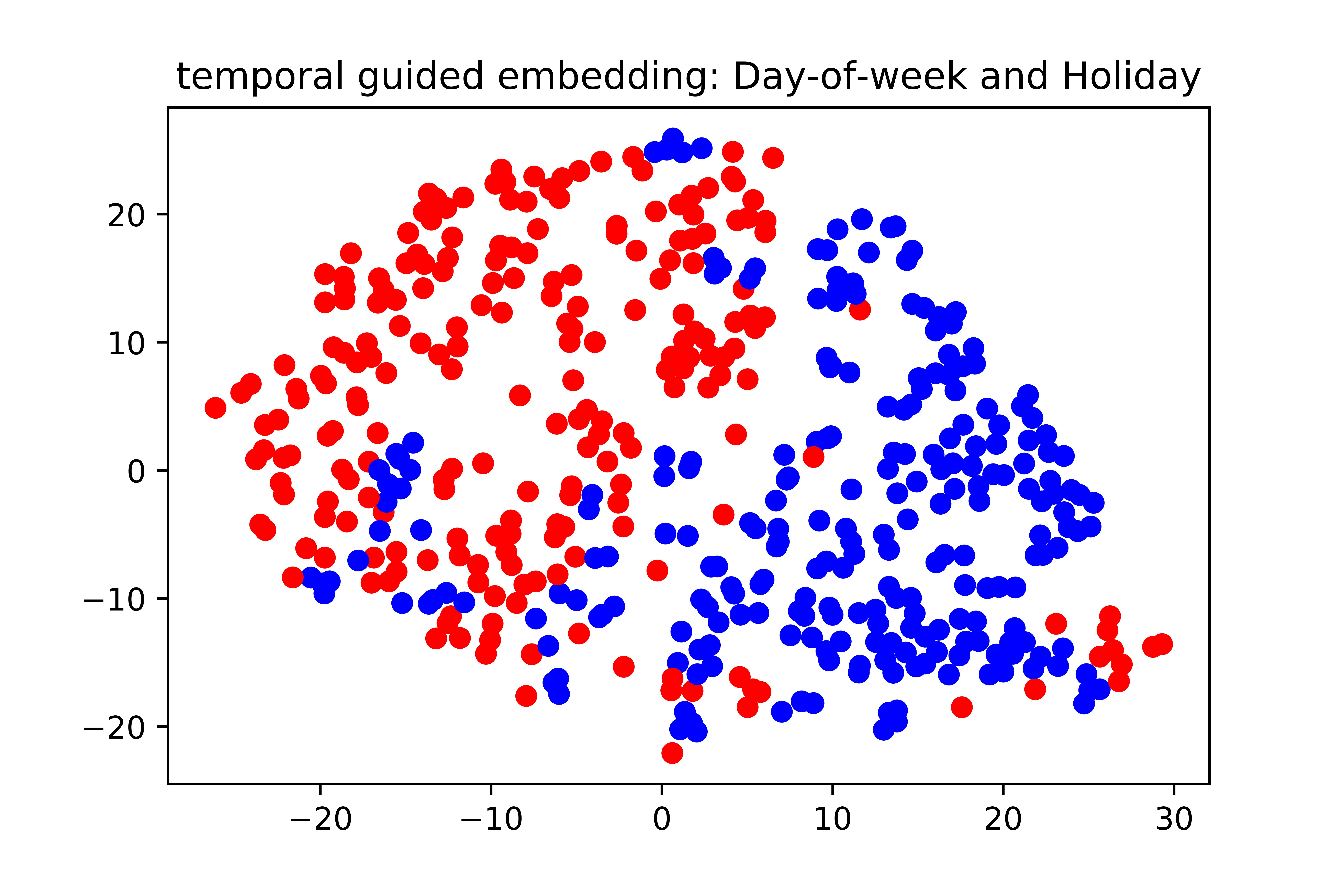}
\caption{t-SNE visualization of temporal-guided embedding of NYC-taxi. Weekday (blue) and holiday \& weekend (red) are clearly classified in the embedding space.}
\label{tge_holi_weekday}
\end{figure}

Although temporal-guided embedding improve forecasting results on all datasets, we can not show meaningful insights on NYC datasets like SEO-taxi.
That is, we find that the adjacent time-of-day vectors tend to be located adjacent to, but it is not obvious to all time-of-day vectors (Figure~\ref{tge_nyc_tod})..
The working day (weekday) and the other days (weekend and holiday) are also divided clearly in the embedding space (Figure~\ref{tge_holi_weekday}).
We conclude that NYC datasets may not have enough number of samples to learn temporal contexts like SEO-taxi, but overall concepts of learning of temporal-guided embedding are similar with large-scale dataset, SEO-taxi.

\subsection{Time-Series Forecasting and Temporal-Guided Embedding}
In this paper, we showed that temporal-guided embedding make forecasting model improve performances and learn temporal contexts explicitly.
The implementation of temporal-guided embedding is simple, but it has theoretical background.
Let an observation of time series is $(x_{t-T+1}, x_{t-T+2}, ..., x_{t-1}, x_{t})$.
From ARMA to recent deep learning models, the forecasting models learn autoregressive model of $x_{t+1}$ with lag $T$ inputs $x_{t}$, ..., $x_{t-T+1}$
\begin{equation}
    p(x_{t+1}|x_{t}, ..., x_{t-T+1})
\end{equation}
where t is time stamp of each data sample.
If a model assume Markov property (not our case), such as LSTM, it becomes
\begin{equation}
    p(x_{t+1}|x_{t}, ..., x_{t-T+1}) = \prod_{i=1}^{T}{p(x_{t-T+i+1}|x_{t-T+i})}.
\end{equation}
TGNet does not assume sequential model and directly learn equation (13).
In general, the model (13) or (14) is corresponded to for all time stamps in training sample, assuming stationary condition.
Because of stationary condition, a model is not feasible when the time series is non-stationary and has different probability distribution according to time stamps $t$.
Thus, Some preprocessings, such as log scaling or differencing, are used to make the series stationary and neural networks effectively make non-stationary series stationary automatically by learning hierarchical nonlinear transformations.
Furthermore, deep learning model contains both model of probability distribution for stationary process (output layer) and preprocessing modules (hidden layers) to make the input stationary.

Note that we can rewrite (13) with random variables of a fixed-length ordered sequence
\begin{equation}
    p(X_{T+1}|X_{T}, X_{T-1}, ..., X_{1})
\end{equation}
where $X_{k}=\left\{ x_t: t \ge k \right\}$ and $(X_{k+1}, X_{k})=\left\{ (x_{t+1}, x_{t}): k \ge i \right\}$.
That is, above equation (13) is a special case when $k=t-T+1$.

We know that equation (15) does not contain any temporal information about specific time $t$, but model probability distribution of input ordered sequence.
That is, it means that the model makes combinations of input values to predict future demand without explicit knowledge or understanding of temporal contexts.
However, the these approach to model time-series is quite different from how human understands time-series, because people learn temporal contexts of data from explicit understanding of time-of-day, day-of-week, or holiday.

Temporal-guided embedding makes the model predict conditional distribution on temporal contexts of forecasting target
\begin{equation}
  p(X_{T+1}|X_{T}, X_{T-1}, ..., X_{1}, \text{TGE}(\tau(X_{T+1})))
\end{equation}
where $\text{TGE}(\tau(X_{T+1}))$ is learned temporal contexts and $\tau(X_{T+1})$ is temporal information vector (time-of-day, day-of-week, holiday, the day before holiday) of random variable $X_{T+1}$.
Temporal-guided embedding explicitly learns temporal contexts of forecasting target and make model extract hidden representations of input sequence conditioned on the embedding.
We replace input of long-term histories from days/weeks ago with temporal-guided embedding and show that the embeddings improve forecasting performances and have interpretable visualizations.
We expect temporal-guided embedding can be used for general time-series modeling in future work.

\section{Atypical Events and Drop-off Volumes}
\begin{figure}[h]
\centering
\includegraphics[width=10cm]{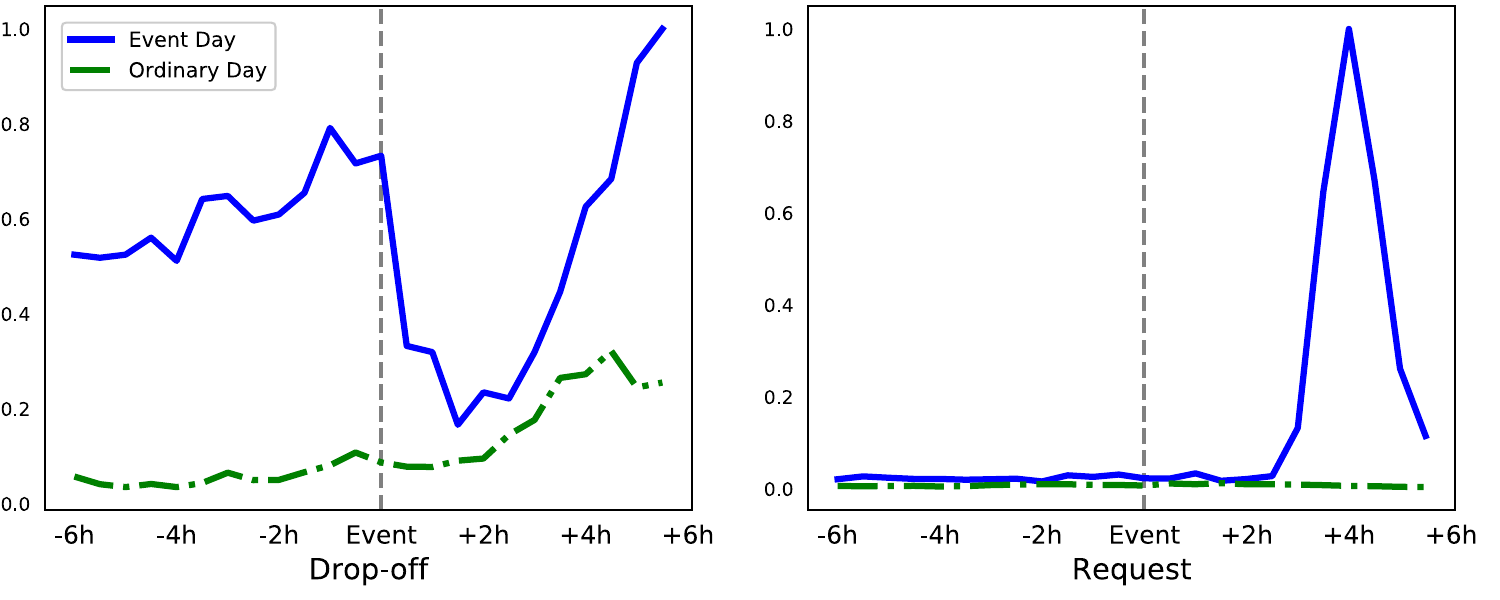}
\caption{Drop-off/pick-up volumes have different patterns when atypical event occurs (blue) from ordinary day (green).}
\label{fig_event_example}
\end{figure}

We conduct evaluation of forecasting performances on atypical event samples, which have extremely large demand volumes, and show drop-off volumes can improve forecasting results.
In fact, we found that the patterns of taxi drop-off at a region were different before atypical events occurred (Figure~\ref{fig_event_example}).
Sudden surge of pick-up requests is observed after atypical events, such as music festival, end and drop-off volumes are much larger than usual before the atypical events start.
Many people rush into the region to participate in the events and the drop-off volumes can be potential demand in future.

\section{Forecasting Performance Details}
The standard deviations with ten repeats are attached in Table~\ref{tab:overall_result_with_std}.
We conclude that our proposed model is significantly competitive to other baseline models.
In the cast of NYC-bike, our model is not significantly better than STDN \cite{yao2018modeling}, but there is no statistically significant difference.
When we think that the number of parameters of TGNet is about 20 times smaller than  STDN and bike demands are vulnerable to weather conditions, we consider our results on NYC-bike promising.
\begin{table*}[h]
\caption {Forecasting results and Standard Deviation with ten repeats on real-world datasets.}
\label{tab:overall_result_with_std}
\centering
\begin{tabular}{l|cccccc}
\toprule
\multirow{2}{*}{Method} & \multicolumn{2}{c}{NYC-bike}            & \multicolumn{2}{c}{NYC-taxi}              & \multicolumn{2}{c}{SEO-taxi} \\
                        & RMSE               & MAPE(\%)            & RMSE                & MAPE(\%)            & RMSE          & MAPE(\%)   \\ \hline
ARIMA                   & 11.53              & 27.82               & 36.53               & 28.51               &    48.92      &     56.43  \\
XGBoost                 & 9.57               & 23.52               & 26.07               & 19.35               &    32.09       &   45.75  \\
STResNet                & 9.80 $\pm$ 0.12          & 25.06 $\pm$ 0.36          & 26.23 $\pm$ 0.33          & 21.13 $\pm$ 0.63          & -             & -  \\
DMVST-Net               & 9.14 $\pm$ 0.13          & 22.20 $\pm$ 0.33          & 25.74 $\pm$ 0.26          & 17.38 $\pm$ 0.46          & -             & -  \\
STDN \cite{yao2018modeling}                   & 8.85 $\pm$ 0.11          & 21.84 $\pm$ 0.36          & 24.10 $\pm$ 0.25          & 16.30 $\pm$ 0.23          & -         \\ \hline
GN                   & 9.09 $\pm$ 0.05 & 22.51 $\pm$ 0.16 & 23.75 $\pm$ 0.30 & 15.43 $\pm$ 0.15 & 28.10 & 37.31 \\
GN + TGE                   & 8.88 $\pm$ 0.09 & 22.37 $\pm$ 0.06 & 22.81 $\pm$ 0.07 & 14.99 $\pm$ 0.07 & 25.96 & \textbf{35.67} \\
TGNet               & 8.84 $\pm$ 0.07 & 21.92 $\pm$ 0.13 & \textbf{22.75 $\pm$ 0.14} & \textbf{14.83 $\pm$ 0.06} & \textbf{25.35} & 35.72 \\
\bottomrule
\end{tabular}
\end{table*}

\end{document}